\documentclass[letterpaper]{article} 
\usepackage[draft]{aaai2027}  
\usepackage[hyphens]{url}  
\usepackage{graphicx} 
\usepackage{natbib}  
\usepackage{caption} 
\usepackage{algorithm}
\usepackage{algorithmic}

\usepackage{newfloat}
\usepackage{listings}
\DeclareCaptionStyle{ruled}{labelfont=normalfont,labelsep=colon,strut=off} 
\floatstyle{ruled}
\newfloat{listing}{tb}{lst}{}
\floatname{listing}{Listing}

\usepackage{booktabs}

\usepackage{amsmath}
\usepackage{amssymb}
\usepackage{comment}

\newcommand{\Agent}{\mathcal{A}}          
\newcommand{\Meta}{\mathcal{M}}           
\newcommand{\Bank}{\mathcal{V}}           
\newcommand{\Pol}{\pi}                     
\newcommand{\Reward}{R}                    
\newcommand{\Task}{x}                      
\newcommand{\Plan}{o}                      
\newcommand{\Cset}{\mathcal{C}}           
\newcommand{\sig}{\sigma}                  
\newcommand{\impl}{\varphi}                

\title{SBCO: Self-Supervised, Verifier-Grounded Harness Optimization \\ For Planning Agents}
\author {
    Vivek Kulkarni\textsuperscript{\rm 1}\corresponding,
    Sudipta Paul,
    Aounon Kumar,
    Nicholas Tzou,
    Srinivas Chappidi
}
\affiliations {
    Samsung Research America\\
    \{v.kulkarni1, sudipta.paul, aounon.kumar, n.tzou, vasu,c\}@samsung.com
}

\begin{document}

\maketitle

\begin{abstract}
Self-improving agents seek to reduce the human engineering  effort behind AI
systems by enabling them to evolve and self-improve their performance over time. Recently, methods like the Darwin G\"odel Machine and the Huxley G\"odel Machine have been proposed which enable open-ended, \emph{recursive} self-improvement through \emph{self-reference} where a coding agent edits its own code. Such self-referential self-improvement methods require that the competence required to perform the task coincides or aligns well with the competence required for self-modification which is the case for coding tasks. For domains or tasks, which do not satisfy the alignment needed, self-referential self-improvement is not available. In such cases, it is possible to adapt the above algorithms to other tasks by removing the self-referential aspect or introducing explicit self-modification of a meta-agent -- both computationally expensive, relying on population or self-modification search over many candidate agents. For planning tasks with explicit constraints, we propose a far cheaper alternative. We introduce \textbf{SBCO}
(\emph{Self-supervised Block Coordinate Optimizer}), a verifier-grounded harness
optimizer in the same closed-loop, improve-from-experience family as the
G\"odel-machine methods, but self-supervised rather than self-referential.
Given an agentic harness, SBCO learns a decomposed bank of verifiers and a harness
policy via approximate block coordinate ascent, improving the agent's outputs from its own
graded feedback---with a fixed meta-agent and no
human labels. Across two domains SBCO \textbf{matches or exceeds}
a customized self-modifying baseline while using \textbf{4--5.5} times less
compute budget.
\end{abstract}


\section{Introduction}
\label{sec:intro}
Building capable AI agents still demands substantial human engineering--hand-crafted prompts, tools, control flow, and checks. \emph{Self-improving} agents aim to reduce that effort by letting a system improve its own problem-solving over time~\cite{hyperagents}. Recently, methods like the Darwin G\"odel Machine \cite{dgm} and the Huxley G\"odel Machine \cite{hgm} have been proposed in this realm. These methods enable open-ended, \emph{recursive} self-improvement of coding agents through
\emph{self-reference} where the coding agent edits its own code to self-improve in a recursive manner. As noted by ~\cite{hyperagents}, the ability to self-improve the improvement mechanism relies on \emph{self-reference} where the competence the agent is scored on  coincides with the competence it needs to improve itself, and gains at the
task translate into gains at self-improving which holds for coding tasks. Beyond coding this alignment generally breaks: improving a planning agent is a coding or specification problem, and a better planner is not thereby a better agent-editor. It is possible to adapt these algorithms to non-coding tasks, by either (a) removing the self-referential nature and introducing a hand-crafted fixed meta-agent (per domain) or (b) introduce self-modification in the meta-agent itself enabling meta-cognition which Hyperagents accomplished ~\cite{hyperagents}. However they also note that introducing meta-cognition (Option b) mainly benefits transfer of the meta-cognitive strategy to other domains with no significant gain in task performance beyond customized fixed meta-agents (Option a). Nevertheless, both options are \emph{computationally expensive}, relying on population self-modification search that generates and evaluates many candidate agents.

We ask whether for planning tasks with explicit constraints, a far cheaper mechanism suffices. We introduce \textbf{SBCO} (\emph{Self-supervised Block
Coordinate Optimizer}), a self-improving verifier-grounded harness optimizer in the same
closed-loop, improve-from-experience family as the G\"odel-machine methods, but
\emph{self-supervised rather than self-referential}. Given an agentic
harness--SBCO learns a \emph{decomposed
bank of per-constraint verifiers} and a \emph{policy} that checks the
agent's output and repairs it when a reliable verifier fires. The optimizer is a
\emph{fixed} meta agent: it improves the harness by approximate block coordinate
ascent with textual gradients, from its own graded feedback and with no human
labels. In summary, our main contributions are:
\begin{itemize}
  \item SBCO, a fixed, verifier-grounded harness optimizer that improves a long horizon planning agent via approximate block coordinate ascent with textual gradients over a learned, decomposed verifier bank and a repair policy.
  \item SBCO reaches the frontier of a customized
    self-modifier at $4$--$5.5\times$ less compute (on par on travel, exceeding on
    shopping).
  \item Our analysis shows that improvement tracks verifier reliability
    ($86\%$ of gains come from verifiers with F1 $\ge 0.80$), and the policy
    transfers across base LLMs.
\end{itemize}

\section{Related Work}
\label{sec:related}
Related work can be categorized into four main areas:
\paragraph{Recursive or Self-referential Self-improvement.} These works propose methods which improve an agent where the agent rewrites its own code \cite{dgm,hgm, stop, zhang2026self}. Because these methods are both self-referential and self-improving, they require that the competence for self-improvement aligns well in the task and this typically is the case for coding.  The variants of the G\"odel machine namely the Darwin G\"odel Machine~\cite{dgm} and Huxley G\"odel Machine~\cite{hgm} belong to this class and propose evolutionary search algorithms which evolves an archive of self-modified agents via mutations and select a final agent from the archive. As mentioned, these methods focus on coding tasks. The works of ~\cite{fernando2023promptbreeder,zhang2026self} are close to that of the Darwin G\"odel Machine but instead only restrict themselves to improving prompts \cite{fernando2023promptbreeder} or adopt a a more bounded search procedure ~\cite{zhang2026self}. 

In contrast to the above which are self-referential and hence by design are limited to primarily modifying only prompts or only coding tasks, SBCO instead shows how one can improve long-horizon planning tasks (breaking away from coding) using an iterative self-improvement procedure using a \emph{fixed meta-agent}. We further distinguish ourselves from ~\cite{hyperagents} in that they introduce meta-cognitive ability to Darwin G\"odel Machine variants and mainly note that this meta-cognitive ability enables transfer of learned cognitive strategy across tasks but does not yield statistically significant improvements over customized Darwin G\"odel Machines. Moreover, because the fundamental nature of open-ended and evolutionary search persists, it can still be computationally expensive to run.  Different from them, our focus is not meta-cognition or transfer of any meta-cognitive strategy but primarily task performance on planning tasks with constraints where we show how to obtain frontier performance with significant compute budget efficiency.

\paragraph{Automated Design of Agents and Workflows.} A parallel line of work uses a fixed
meta agent to search over agent \emph{designs}---inventing or composing agentic
building blocks (chain-of-thought, debate, tool use, ensembling) into agents or
workflows under aggregate benchmark reward: ADAS \citep{hu2025automated} invents new agent programs, AFlow \citep{zhang2025aflow} searches code-represented workflows, and \citet{multiagentdesign} optimize multi-agent prompts and topologies. The product is a new, often multi-agent, system. SBCO does not yield focus on multi-agent system design and instead grounds optimization in a
\emph{learned, decomposed bank of per-constraint verifiers} plus a repair
policy that internally utilizes a correctness signal per constraint as well as an overall reward score. 

\paragraph{Harness Optimization.} A broad line of work improves a base model by
optimizing the \emph{harness} wrapped around it rather than redesigning the agent.
Some approaches update model parameters while most are training-free. STaR \citep{zelikman2022star} fine-tunes on self-generated rationales filtered by correctness. Self-Refine \citep{madaan2023self} refines the output of a LLM by just asking a critique model to provide feedback on that output which it would add to the LLM and iteratively refine the output (essentially focusing on on only prompt refinement). AlphaEvolve \citep{novikov2025alphaevolve} focuses on evolving entire algorithms and programs with the aim of enabling scientific discovery using evolutionary approaches. More recent harness tuning approaches like Trace2Skill learn a skill-bank over time ~\cite{ni2026trace2skill}, Meta-Harness \citep{lee2026meta} tune a harness \emph{monolithic} object against a scalar reward, AutoHarness \citep{lou2026autoharness} \emph{tree-searches} an LLM-coded harness. While SBCO shares the ``training-free'' aspect with these approaches, it differs from these approaches in several ways: (a) it optimizes the entire harness and not just prompts or skills (b) it does not model the harness as a mono-lith but adopts a structured optimization approach (c) adopts an iterative approach instead of a tree-search and (d) focuses on long horizon planning tasks with constraints.

\paragraph{Using Verifiers to Improve Task Performance} There has been some prior work in using verifiers (either using an LLM as judge or deterministic functions) to improve task performance of LLMs. The first line of work uses them to provide a training signal, especially in reinforcement learning based training by introducing process reward models that score intermediate reasoning steps \citep{setlur2025rewarding,khalifa2025process,pronesti2026beyond, zhang2026agentv}. The second line, closest to ours, applies verifiers over a fixed generator
\citep{arora2023learning,zhang2026verified,pezeshkpour2026autopyverifier}. \citet{arora2023learning} train a verifier to detect constraint violations in a blocks-world action space and re-plan when a violating action is found; they do not optimize the harness, only proposing a replan-if-failed loop. \citet{zhang2026verified} address complex question answering with a plan--execute--verify--replan framework that decomposes a question into sub-questions, solves each with a sub-agent, and verifies sub-answers before composing a final answer but the verifier is a fixed, hand-designed
LLM-as-judge prompt. Finally, \citet{pezeshkpour2026autopyverifier} propose a search-based method that infers both a set of criteria and corresponding verifiers for judging LLM outputs, but do not learn a repair workflow. In contrast, SBCO jointly learns to optimize both the
verifier set and the underlying harness (replan strategy) in a self-supervised manner targeting long-horizon planning tasks.
\section{SBCO}
\label{sec:method}
\subsection{Overview}
\label{sec:setup}
We are given a \emph{base agent} $\Agent$ that performs a task $\mathcal{X}$ and produces a plan $\mathcal{O}$.  We assume we can evaluate any generated plan assuming a black-box evaluation suite that outputs a concrete reward (score) and reasonable descriptions of errors (eg. plan exceeded budget). We also assume some reasonable knowledge or description of the semantics of constraints (eg. budget exceeded means the total cost of the itinerary travel, accommodation exceeds the user specified budget). 

\begin{figure}
    \centering
    \includegraphics[width=\linewidth]{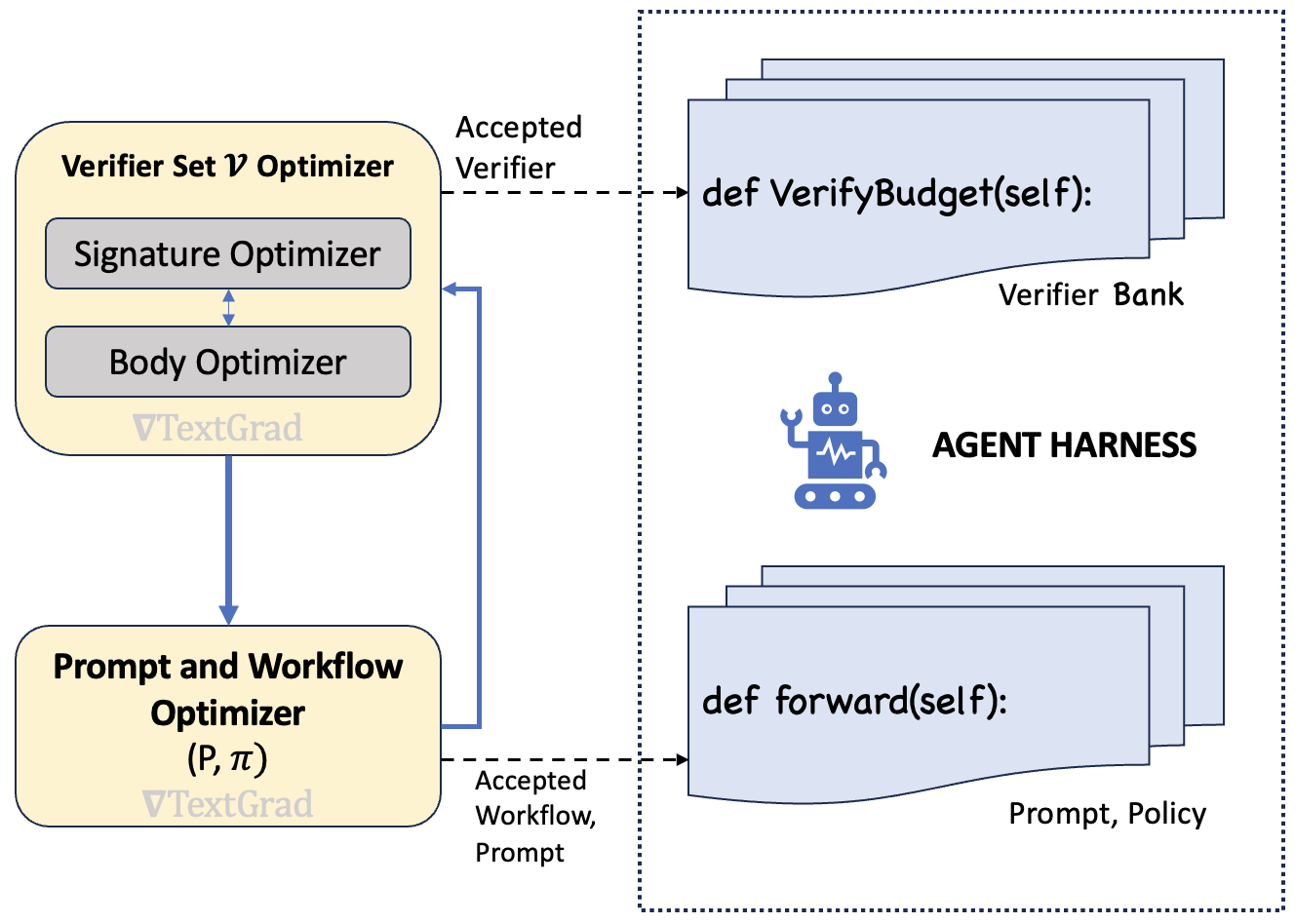}
    \caption{SBCO iteratively alternates between two main phases: (a) Optimizing the verifier set $\mathcal{V}$ fixing $\mathcal{P}$, $\Pi$ and (b) Optimizing the prompt $\mathcal{P}$ and the policy $\Pi$ fixing $\mathcal{V}$.}
    \label{fig:placeholder}
\end{figure}

Because SBCO optimizes in structured manner, let us formalize the components of the agentic harness. Let us define the agentic harness to be a tuple of the form $(\mathcal{P, T, W})$, a prompt $\mathcal{P}$, a set of frozen tools $\mathcal{T}$ that the LLM can use to perform the planning task, and $\mathcal{W}$ the agentic work-flow. We further partition $\mathcal{W}$ into two parts: (a) $\mathcal{V}$: a set of verifiers we will learn/introduce and (b) $\Pi$ - the actual code or policy that will be executed to perform the task. SBCO seeks to maximize the expected reward $\mathbb{E}_{\Task}[\Reward(\Task,\Plan);\mathcal{P, T, W}]$ over the task distribution by optimizing the following components of the harness: $\mathcal{P}$, $\mathcal{V}$, the set of verifiers and $\Pi$. During optimizations SBCO iteratively alternates between two main phases: (a) Optimizing the verifier set $\mathcal{V}$ fixing $\mathcal{P}$, $\Pi$ and (b) Optimizing the prompt $\mathcal{P}$ and the policy $\Pi$ fixing $\mathcal{V}$ -- both optimizations done in textual space using textual gradients obtained through a critic/strategist (see Figure \ref{fig:placeholder}). In following sections, we will describe each phase in detail. 

\subsection{Optimizing the Verifier Set $\mathcal{V}$}
\label{sec:verifier}
For each constraint $c$ we seek to learn a verifier that can check whether $c$ is satisfied or not.  Learning such a verifier involves two coupled blocks (a) \textbf{Signature Block}: A function signature that describes the input and output signatures, and the semantics of the verifier. (b) \textbf{Implementation Block}: this is the actual implementation or the body of the function/verifier. To judge the quality of a learned verifier, we use a small set of evaluations from a baseline run which allows us to construct a ground-truth set (positive examples and negative examples corresponding to the specific constraint passing or failing) and use a precision/recall (F1) score to ascertain how good the verifier is for that specific constraint. This provides a very cheap and computationally inexpensive way to judge verifier quality.  We optimize both blocks using textual gradients by keeping one block fixed while optimizing the other. 
\paragraph{Quality Checks and Acceptance Gating the Verifier Set} Because learning verifiers for some constraints may be hard, to ensure a very poor verifier does not enter the final bank, we introduce quality checks and acceptance gating.  A verifier is \emph{accepted} and
entered into the bank iff it clears a reliability floor (a precision floor $\tau_P=0.80$  and recall floor $\tau_R=0.55$),
Verifiers that cannot clear the reliability floor because the needed signal is not
extractable from the available fields, or precision/recall stays below
threshold are \emph{left out} of the bank. This ensures the bank meets at-least a certain level of base quality. 

\subsection{Optimizing the Prompt $\mathcal{P}$ and the policy $\Pi$}
\label{sec:policy}
In this phase, we keep the learned verifier set $\mathcal{V}$ fixed, and only optimize the prompt $\mathcal{P}$, and the code (policy) $\Pi$. Importantly, policy $\Pi$ can decide how to use the set of verifiers to improve task performance (eg. check plans and repair them as needed). It is worth noting that in practice $\Pi$ can be completely free-form code or may be constrained to follow some enforced structure -- an implementation choice. Once again, as in prior stage, we optimize these using textual gradients ~\cite{yuksekgonul2024textgrad} using the reward returned by the evaluation benchmark. Finally in both phases, we introduce a keep-best ratchet to ensure we do not regress to a worse setting when optimizing each block. 

\subsection{SBCO -- Outer Loop}
\label{sec:bca}
The outer loop of SBCO mainly iterates through every block in sequenced-order (encoding dependencies) and optimizes each block keeping the rest of the blocks fixed (Alg.~\ref{alg:sbco}). The inner optimization routine (Alg.~\ref{alg:block}) which optimizes each block performs tactical refinement interleaved with occasional strategic pivots (to break out of poor regimes). 

Specifically, the tactical refinement phase uses textual gradients to optimize the block. To break out of local optima,  we introduce the ability to strategically pivot if the improvement over several iterations does not meet the acceptance gate. During such a strategical pivot the critic sees all prior strategies tried, their failures and proposes a completely new strategy (try a different tool, a different algorithm etc) or declares a potential information ceiling. 

Furthermore, because the verifier set optimization uses a surrogate reward metric for optimization that is not exactly the same as the final task reward objective, it is possible that the policy block optimizer would like to flag potential mismatches back to the verifier optimization phase to re-visit/re-open an already accepted constraint verifier. Therefore, we enable down-stream phases to potentially escalate to upstream phases to re-open upstream artifacts.  

Finally, given access to the task agent, the evaluation  benchmark and its output, the loop is a closed loop -- a loop consisting of 
propose~$\to$~grade~$\to$~critique~$\to$~revise, steps closed over the model's own
outputs. This is the sense in which SBCO is \emph{self-improving}---the loop
improves its artifacts from its own experience---while remaining
\emph{self-supervised} but not recursive/self-referential.

\begin{algorithm}[t]
\caption{SBCO (outer loop)}
\label{alg:sbco}
\begin{algorithmic}[1]
\REQUIRE agent $\Agent$; constraints $\Cset$; gate $(\tau_P,\tau_R)$; epochs $E$
\ENSURE verifier bank $\Bank$, policy $\Pol$
\STATE $\Bank \gets \emptyset$; initialise per-constraint state; $\Pol \gets \Pol_{\text{seed}}$ \COMMENT{no-op seed}
\FOR{$e=1$ to $E$}
  \FORALL{active $c \in \Cset$}
     \STATE $\sig_c \gets \textsc{BlockOptimize}(\textsf{signature}, c)$
     \STATE $\impl_c \gets \textsc{BlockOptimize}(\textsf{implementation}, c \mid \sig_c)$
     \IF{$\text{accept}(\impl_c)$}
        \STATE $\Bank \gets \Bank \cup \{\textsc{compile}(\impl_c)\}$ \COMMENT{reliability gate}
     \ENDIF
  \ENDFOR
  \STATE $\Pol \gets \textsc{BlockOptimize}(\textsf{policy})$ \COMMENT{policy block}
  \STATE \textsc{Escalate}$(\Pol, \Bank)$ if needed;\ 
\ENDFOR
\STATE \textbf{return} keep-best $\Bank, \Pol$
\end{algorithmic}
\end{algorithm}

\begin{algorithm}[t]
\caption{\textsc{BlockOptimize}: tactical refinement + strategic pivot (fixed $\Meta$)}
\label{alg:block}
\begin{algorithmic}[1]
\REQUIRE block type; unit (constraint or bank); tactical budget $T$; max pivots $\rho$
\STATE $b^\star \gets \text{seed}$;\ $s \gets \text{seed}$
\FOR{$p=0$ to $\rho$ \textbf{(outer: strategic pivots)}}
  \IF{$p > 0$}
     \STATE $s \gets \textsc{Strategist}(\text{brief})$ \COMMENT{pivot: structurally new, or InfoCeiling}
  \ENDIF
  \FOR{$t=1$ to $T$ \textbf{(inner: tactical refinement)}}
     \STATE $b \gets \textsc{Generate}(s)$;\ $g \gets \textsc{Grade}(b)$
     \IF{$\textsc{Better}(g, g(b^\star))$} \STATE $b^\star \gets b$ \COMMENT{keep-best} \ENDIF
     \IF{$\textsc{MeetsGate}(g)$} \STATE \textbf{return} $b^\star$ \ENDIF
     \STATE $s \gets \textsc{TextGradStep}(\text{critic}, b, \textsc{GradeReport}(g))$
  \ENDFOR
  \IF{$\text{accept}(b^\star)$ \OR $\lnot\,\textsc{ShouldPivot}()$}
     \STATE \textbf{break} \COMMENT{converged, or strategist declines to pivot}
  \ENDIF
\ENDFOR
\STATE \textbf{return} $b^\star$
\end{algorithmic}
\end{algorithm}

\section{Experiments}
\label{sec:setup-exp}

\paragraph{Benchmarks and Evaluation Metrics} To evaluate our method, we consider the popular \emph{DeepPlanning} benchmark ~\cite{zhang2026deepplanning} which consists of two long-horizon planning tasks with constraints: (a) \textbf{Travel Planning} -- where the task agent is provided access to task-specific tools (that query a local database) and is tasked to generate a multi-day itinerary based on the specification of the user. The user specifies their travel requirements and outlines constraints (eg. budget should be within $1000$ dollars, need to stay in a 3.5 star hotel, a plan for 4 days and 3 nights.).  (b) \textbf{Shopping} -- where the task agent is provided access to a set of tools (that query a product catalog) and is asked to create a cart consisting of items from a product catalog that meet the user's request. The user's request specifies what kind of items they want and potential constraints (eg. need a 4.5 star rated Ralph Lauren sweater). Both tasks can be challenging because the agent needs to reason about implicit constraints (like size based on user profile) and many explicit constraints. We evaluate our method on multiple dimensions:
\begin{itemize}
\item \textbf{Task Performance}: We report standard task performance metrics as specified by the benchmark -- namely Composite score and Case Accuracy for Travel Planning, and Match Score and Case Accuracy for Shopping. 
\item \textbf{Compute Budget}: To evaluate the compute efficiency of our algorithm against baselines like HGM, we use the same protocol they used to quantify the compute budget -- namely the number of plans generated during the optimization process (which they term as evaluation budget).
\item \textbf{Latency}: We also report the average latency of methods to generate a plan. 
\end{itemize}

\paragraph{Baseline Methods} We compare against state-of-the-art
\emph{self-improving} methods that, like SBCO, drive improvement with a
\emph{fixed} meta agent--a handcrafted, \emph{non-metacognitive} improvement procedure, following the terminology of \citet{hyperagents}: (a) the Huxley-G\"odel Machine self-improving algorithm HGM~\citep{hgm} and a domain-customized variant, HGM-C. Note we consider HGM (and its customization) over DGM~\citep{dgm} because HGM is known to already outperform DGM on task performance and budget. We also report a no-op \emph{Baseline} (the base agent with no improvement loop). We deliberately omit
the recent \emph{metacognitive} self-modifier DGM-H~\citep{hyperagents} for two reasons grounded in that work itself. 
First, \citet{hyperagents} note that their proposed meta-cognitive self-modification (DGM-H) yields \emph{no statistically significant gain} in \emph{task performance} over the fixed-meta customized baseline DGM-C (a higher median
but $p>0.05$); its measured benefit is actually cross-domain \emph{transfer} and
accumulation, which we do not claim. 
Because we make only a task-performance claim and make no claims about \emph{meta-cognitive} ability or cross-domain transfer ability, the appropriate state-of-the-art comparison for task-performance is therefore the strongest fixed-meta self-improving baselines, HGM and HGM-C.
\paragraph{Evaluation Protocol.} The task-agent use GPT-5.4-mini as the backbone LLM while the critic/strategist are GPT-5.4-medium unless specified. For HGM baseline, we set the evalauation budget to $8000$ with all other parameters set to their default. For SBCO, we set the total number of epochs $E=3$ for the outer-loop, and for the block-optimizer, the maximum number of iterations for tactical refinement is set to $5$ with at-most $2$ strategic pivots. SBCO uses a subset of the benchmark data for training (and validation). However, \emph{final performance} is reported on the full task set (training
$+$ validation). This matches the reporting protocol of the G\"odel-machine baselines (DGM and HGM) which
select and report on the same benchmark rather than a held-out test split, so all
methods in Table~\ref{tab:main} are evaluated identically and the comparison is
apples-to-apples. All numbers are the mean of two runs of the policy to account for variance in planning \footnote{The number of independent runs of the policy was the best choice under the budget compute we had available.}. 

We argue that this setting does not confer an unfair advantage here,
for the same reason it does not for DGM or HGM: improvement is mediated by
\emph{general artifacts}, not by memorizing per-task answers. SBCO's verifiers are
constraint-checking \emph{programs} (e.g.\ ``is the chosen hotel the cheapest of
the required brand?'') and its policy is general repair logic; neither of which encodes
a instance specific answers. This is similar to DGM/HGM where the improvement comes from general code and not per-instance solution look-up. To alleviate concerns of SBCO overfitting, we explicitly also report separately train and validation set performance of the best policies and show the train-validation gap is small. Further-more we also show that the learned artifacts generalize directly to different LLM backbones of the task agent further underscoring the generalizability of SBCO.

\section{Results}
\label{sec:results}
\subsection{Quantitative Results}
Table~\ref{tab:main} reports the main comparison on both domains. Note that budget is the
number of plans generated (tasks evaluated) during the optimization phase of each algorithm. First, observe that the naive Huxley G\"odel-machine method fails to improve meaningfully despite $8\times$ the budget (on Shopping it does not move from the baseline at $8000$ plans). Second the customized HGM (HGM-C) (which includes a hand-crafted meta-agent prompt and guidelines) yields significant improvement over the baseline ($83$ vs $76$ on Travel, and $91$ vs $83$ on Shopping).  This performance however used a budget of $4000$ plan generations. On the other hand, SBCO reaches the quality frontier of HGM-C at $4$--$5.5\times$ lower budget. On Travel it matches HGM-C ($84$ vs.\ $83$ composite, matching case accuracy) and on Shopping it exceeds it ($+3$
match, $+9$ case accuracy). Finally, to assess any potential over-fitting of SBCO to training samples, we compare SBCO's performance of the final harness on on the training and validation splits and compute the train-validation gap.  Across both domains the train–validation gap is negligible or favors validation — Travel 84.3 (training) vs. 83.9 (validation) and Shopping 93.7 (training) vs. 94.8 (validation) — indicating SBCO does not over-fit to the data it selects on. Finally, because self-improving methods learn an improved policy, we incur a cost of increased average latency for plan generation as is expected compared to the baseline. 

\begin{table*}[t]
\centering
\begin{tabular}{l cccc cccc}
\toprule
 & \multicolumn{4}{c}{\textbf{Travel}} & \multicolumn{4}{c}{\textbf{Shopping}} \\
\cmidrule(lr){2-5}\cmidrule(lr){6-9}
Method & Composite & Case Acc. & Latency (s) & Budget & Match & Case Acc. & Latency (s) & Budget \\
\midrule
Baseline             & 76          & 28 & 414 & -           & 83          & 50          & 106 & - \\
HGM          & 77          & 30 & 366 & 8000          & 83          & 50          & 106 & 8000 \\
HGM-C                & 83          & 30 & 715 & 4000          & 91          & 70          & 245 & 4000 \\
\textbf{SBCO (ours)} & \textbf{84} & 30 & 523 & \textbf{733}  & \textbf{94} & \textbf{79} & 132 & \textbf{989} \\
\bottomrule
\end{tabular}
\caption{SBCO matches or exceeds HGM-C at $4$--$5.5\times$ lower
budget; the naive self-modifier barely improves at $8\times$ the budget. Budget $=$ number of plans generated in optimization phase. Quality is composite
score (Travel) / match score (Shopping) on a $0$--$100$ scale (scores rounded to $2$ significant digits); higher is better
for all columns except Latency and Budget, where lower is better. \textbf{Bold}
$=$ best among methods.}
\label{tab:main}
\end{table*}

\paragraph{Generalization to weaker models} Having empirically demonstrated that SBCO matches the quality frontier of strong baselines (and even outperforms in some cases) at a fraction of the compute budget, we now shift gears to assess the generalization of SBCO to a weaker model in two settings: (a) Policy Transfer -- Do the verifiers and learned policy learned on a model transfer to a different model (potentially weaker)?  (b) Does SBCO also help with a different weaker LLM used as the backbone when it is optimized from scratch?. 

To assess policy transfer, we consider the Travel task and use the best policy and verifiers obtained with GPT-5.4-mini as the  backbone task-agent LLM and use them \emph{as is} on a task-agent powered by Mistral model, the results of which we show in Table~\ref{tab:xmodel}. Note that indeed we are able to still lift the performance of the weaker agent suggesting that our learned verifiers and policy do not overfit to the underlying back-bone LLM. 
 
To assess whether SBCO itself generalizes to task-agents with different back-bone LLMs, we  instantiate a travel agent with Mistral as the backbone LLM and optimize the harness from scratch, the results of which are presented in  Table~\ref{tab:weakagent}. We note that Mistral is a substantially weaker planner (baseline composite $32.2$ vs.\ GPT's
$76.5$). SBCO nonetheless lifts it by $+9.6$ composite
(Table~\ref{tab:weakagent}), comparable to the $+9.9$ obtained by transferring
the GPT-learned policy. The absolute ceiling, however, remains bounded by the base agent: SBCO raises Mistral to $41.8$ but cannot approach GPT's $84.3$. Verify-and-repair recovers
violations the agent \emph{can} fix once flagged; it does not elevate the base planning
capability of the agent which is considerably weaker than GPT-5.4-mini. This experiment thus demonstrates the robustness of our learned verifiers, policy to different backbone LLMs of the task agent. 

\begin{table}[t]
\centering
\begin{tabular}{ll ccc}
\toprule
\textbf{Domain} & \textbf{Task agent} & \textbf{Base} & $+$\textbf{SBCO} & $\Delta$ \\
\midrule
Travel  & GPT-5.4-mini & 76.5 & 84.3 & $+7.8$ \\
Travel & Mistral      & 32.2 & 42.1 & $+9.9$ \\
Shopping & GPT-5.4-mini & 83.4 & 94.6 & $+11.2$ \\
Shopping & Mistral      & 62.3 & 94.4 & $+32.1$ \\
\bottomrule
\end{tabular}
\caption{Cross-model transfer: a learned policy (where task agent was GPT-5.4-mini) applied as-is to a
Mistral task agent, showing policy and verifiers transfer to other backbone LLMs for the task agent.}
\label{tab:xmodel}
\end{table}

\begin{table}[t]
\centering
\begin{tabular}{lccc}
\toprule
\textbf{Task agent (Travel)} & \textbf{Base} & $+$\textbf{SBCO} & $\Delta$ \\
\midrule
GPT-5.4-mini (from scratch)  & 76.5 & 84.3 & $+7.8$ \\
Mistral (from scratch)       & 32.2 & 41.8 & $+9.6$ \\
Mistral (policy transfer)    & 32.2 & 42.1 & $+9.9$ \\
\bottomrule
\end{tabular}
\caption{SBCO on a weaker task agent (Travel). SBCO run end-to-end
against a weaker agent (Mistral) matches the lift of transferring a strong
agent's policy; the optimizer is identical in all rows, only the base agent
differs.}
\label{tab:weakagent}
\end{table}

\subsection{Qualitative Results and Analysis}
We now turn to qualitatively analyzing the learned verifiers, policies to gain insights into functionality learned by the improved agents.
\paragraph{Reliability of the learned verifiers} For both tasks, we observe that the learned verifiers are generally highly reliable with many verifiers having near-perfect precision (mean precision $>0.97$ for both tasks). Ranking verifiers by F1 within each domain reveals insights into which constraints are easy for SBCO to learn to verify and which constraints it still finds challenging.

\emph{Travel.} From Table~\ref{tab:travel-verifiers} we observe that SBCO is very good at learning verifiers for constraints that involve matching on a specific attribute (hotel rating), a clear top ranking (cheapest direct) or a very clear mathematical constraint (plan within budget) However we note that it is still challenging to learn verifiers for constraints that involve geo-proximity look-up and slightly implicit constraints (where there may be grader-specific tolerances eg. \emph{closest} or \emph{cheapest} restaurant near an
attraction, reasonable transfer time). We hypothesize that this is because the feedback signal in the evaluation log for these constraints is not rich enough -- for eg. what time buffer is deemed to be ``reasonable in transfer time'' or its challenging for the model to know how exactly to compute geo-proximity for restaurants and near-by attractions since this likely requires complex queries and ranking. We also note that by class, personalized (hard) constraints fare slightly
better than commonsense ones 
because most hard constraints typically involve clean lookups and
rankings though we note the hard class appears to be bimodal, with the best \emph{and} worst travel
verifiers both belong to the hard class (perfect lookups vs.\ completely failed geo-spatial checks).

\emph{Shopping.} Similar conclusions hold for Shopping where SBCO is able to learn reliable verifiers for checking whether a attribute matches, or perform well-defined combinatorial combination checks (eg. whether a required product filter (color, brand, size, delivery time, rating, review count or threshold, sales volume) is satisfied, all at precision $1.00$ (F1 up to $1.00$). It also learns the combinatorial cheapest-combination check (F1 $0.83$). It generally struggles on constraints that involve over-inclusion where we note that the verifier over-flags and does not meet the acceptance gates.

\begin{table}[t]
\centering
\begin{tabular}{llccc}
\toprule
\textbf{Constraint} & \textbf{Class} & \textbf{P} & \textbf{R} & \textbf{F1} \\
\midrule
\multicolumn{5}{c}{\textit{Top 5 (best F1)}}\\
\midrule
hotel\_highest\_rated        & H & 1.00 & 1.00 & 1.00 \\
train\_cheapest\_direct      & H & 1.00 & 1.00 & 1.00 \\
budget\_constraint           & H & 1.00 & 1.00 & 1.00 \\
restaurant\_must\_eat\  & H & 1.00 & 1.00 & 1.00 \\
diverse\_attrn\_options & CS   & 1.00 & 1.00 & 1.00 \\
\midrule
\multicolumn{5}{c}{\textit{Bottom 5 (worst F1)}}\\
\midrule
restaurant\_closest\_to\_attrn     & H & 0.00 & 0.00 & 0.00 \\
restaurant\_chpest\_nr\_attrn & H & 0.00 & 0.00 & 0.00 \\
reasonable\_txfer\_time               & CS   & 1.00 & 0.17 & 0.29 \\
flight\_shrtst\_time\_direct       & H & 1.00 & 0.25 & 0.40 \\
restaurant\_spec\_tag\_near        & H & 0.26 & 1.00 & 0.42 \\
\bottomrule
\end{tabular}
\caption{Travel: Top $5$ best and worst learned verifiers by F1. \emph{Top}: 5
constraints  F1$=$1.00 (local, decidable predicates), spanning check
types. \emph{Bottom}: the 5 lowest, clustering in geo-spatial proximity
(F1$\approx$0; no field encodes inter-entity distance) and derived-quantity
reasoning (temporal windows), and exhibiting all three error
modes---no signal (P$=$R$=$0), low recall (P$=$1.00), and over-flagging (low P).
Class: H $=$ personalized hard, CS $=$ commonsense.}
\label{tab:travel-verifiers}
\end{table}

\begin{table}[t]
\centering
\begin{tabular}{llccc}
\toprule
\textbf{Constraint} & \textbf{Kind} & \textbf{P} & \textbf{R} & \textbf{F1} \\
\midrule
\multicolumn{5}{c}{\textit{Top 5 (best F1)}}\\
\midrule
sales\_volume@L1  & miss   & 1.00 & 1.00 & 1.00 \\
color\_filter@L1  & miss   & 1.00 & 1.00 & 1.00 \\
review\_count@L1  & miss   & 1.00 & 0.86 & 0.92 \\
delivery\_time@L1 & miss   & 1.00 & 0.86 & 0.92 \\
coupon@L3         & coupon & 1.00 & 0.86 & 0.92 \\
\midrule
\multicolumn{5}{c}{\textit{Bottom 5 (worst F1)}}\\
\midrule
extra\_product@L3 & extra & 0.14 & 0.71 & 0.23 \\
extra\_product@L1 & extra & 0.33 & 0.78 & 0.46 \\
extra\_product@L2 & extra & 0.53 & 0.88 & 0.66 \\
season\_filter@L1 & miss  & 1.00 & 0.57 & 0.73 \\
size\_filter@L1   & miss  & 1.00 & 0.60 & 0.75 \\
\bottomrule
\end{tabular}
\caption{Shopping: best and worst verifiers by F1. \emph{Top}:
missing-required-attribute checks (\emph{miss}) and coupon presence, all at
precision $1.00$. \emph{Bottom}: over-inclusion (\emph{extra}) checks, which fail
on \emph{precision} ($0.14$--$0.53$), plus two missing-attribute checks that have low \emph{recall}.
Kind: miss $=$ missing required attribute, extra $=$ over-inclusion.}
\label{tab:shopping-verifiers}
\end{table}

\paragraph{Policy analysis of the improved agents.} We analyze the improved policies of the task agent. Broadly, we observed that the learned policies are not generic ``re-plan on failure'' heuristics but repair strategies thar are adapted to the specific task and agent capability. 

\emph{Travel} learns a gated, LLM-replan-centric pipeline with the following main components:
\begin{itemize}
  \item \textbf{Tiered reliability gating.} It acts on hard verifiers only at
    precision $\ge 0.98$ and on commonsense verifiers only at $\ge 0.999$, and
    only for four common sense constraints below:
    (\texttt{traceable\_accommodation}, \texttt{diverse\_attraction\_options},
    \texttt{seamless\_intercity\_transfers}, \texttt{validated\_meals});
    everything else is ignored.
  \item \textbf{Three-stage repair.} A deterministic code patch for one
    structural check (blank the final-day accommodation for
    \texttt{traceable\_accommodation}) which can be done very reliably in code; then a \emph{single batched} LLM replan
    that lists all trusted-hard failures \emph{and} the passing checks to
    preserve, instructing the agent to copy exact literals verbatim (names,
    flight/train numbers, dates); then a narrow commonsense re-plan only if no hard failures remain.
  \item \textbf{Conservative verify-and-accept.} Every repair re-runs the
    verifiers and is kept only if it \emph{strictly} reduces the trusted-fail set
    with no regression, otherwise it reverts to the pre-repair plan (with at most two
    replans).
\end{itemize}

\emph{Shopping} -- In contrast with Travel where the replan strategy was mainly accomplished by calling an LLM, for Shopping the repair-policy is code-application-dominant and has the following main components:
\begin{itemize}
  \item \textbf{Near Deterministic Plan Repair via Code} Because many of the verifiers in order to verify the constraint end also end up knowing what the correct products are (as a side-effect), the policy prefers to use that information, but crucially realizes that it is possible to make changes to the cart reliably using code avoiding LLM calls for the most part. So the policy is near-deterministic and uses code to modify the LLM generated cart. 
  \item \textbf{Append to cart, do not remove existing products in cart.} The policy only adds missing products/coupons and chooses to not delete existing ones. This highlights  a potential issue with the standard reward function for Shopping. Based on definition of the match score and observing that extra items do not affect the match score, the optimizer realized that the reward essentially is a recall-only metric where the final reward does not penalize extra items. It uncovers this potential ``hole'' and and uses this to maximize the reward. That said, note that this ``gap'' was uncovered by the optimization agent itself and reveals the need for better reward metrics for that benchmark.  
  \item \textbf{Try an LLM replan after code-check repair if needed} For failures remaining after the code append, it issues at most two grouped replans (failures grouped by sub-query, exact ids named, ``preserve all existing, add cheapest valid missing''), then re-merges the returned cart in code and runs a deterministic append pass. 
\end{itemize}

\paragraph{Error Analysis} Finally, we conduct a error analysis of our deployed policy to identify what are the top drivers of the improvement in task performance and identify what error classes are still challenging. Comparing the deployed policy to the
baseline agent on the gold grader shows that the errors SBCO fixes are precisely
those it can verify reliably. 

For Travel, the largest per-constraint gains
over baseline come from a handful of banked, high-precision verifiers (we compute fix and
false-repair counts averaged across two evaluation runs and rounded; a ``fix'' is
a single constraint flipping fail$\to$pass within a task, not a fully solved
instance). The policy yielded the greatest improvement for
\texttt{seamless intercity transfers} (6 fixes, 1 false repair, net gain $+5$),
\texttt{hotel star service required} and \texttt{train departure time range}
(4 fixes, net gain $+4$ each), and \texttt{diverse attraction options} (3 fixes,
net gain ${\approx}{+}2$). While \texttt{reasonable transfer time, cost calculation correctness} remain among the top dominant errors still  persisting, we saw marginal gains on these as a side effect where fixing a different constraint incidentally corrects these (net gain ${\approx}{+}2$ and $+1$ respectively).

On Shopping, SBCO lifts the score from $0.828$ to $0.940$
($+0.112$), but the gain is uneven across difficulty levels. Most of it comes
from L1 and L2: adding missing single-attribute products (L1, $+0.096$) and
repairing cheapest-within-budget combinations (L2, $+0.154$---the largest
per-level gain), both of which SBCO verifies and repairs reliably. L3 improves
least ($+0.045$) because it requires coupon-stacking optimization, for which
SBCO does not learn a reliable verifier; its gains are therefore limited here on this level.

\section{Limitations}
\label{sec:limitations}
Although we have demonstrated the effectiveness of SBCO empirically both quantitatively and qualitatively, our work is not without limitations. First,  we focus and evaluate on two constraint-based planning domains. Whether SBCO's advantages extend to domains without explicit, checkable constraints---or to tasks where verification is as hard as solving---remains open. Second, like all training free methods, the gains offered by our method will be limited by the power of the base  LLM.  Finally, SBCO improves a harness
with a \emph{fixed} meta agent; it does not have meta-cognitive abilitiies like that of ~\cite{hyperagents}. Extending the loop to enable  meta-cognitive abilities is left to future work.

\section{Conclusion}
\label{sec:conclusion}

We introduced \textbf{SBCO}, a self-improving, verifier-grounded harness
optimizer for planning agents. Where the G\"odel-machine lineage achieves
self-improvement through evolutionary search which is computationally expensive, SBCO offers a cheaper alternative for long horizon planning tasks with constraints. SBCO  learns a  bank of decomposed verifiers and a repair policy, optimized by a
\emph{fixed} meta agent via approximate block coordinate ascent with textual
gradients. It thus belongs to the same closed-loop, improve-from-experience family
as those methods, but is self-supervised rather than self-referential. On two constraint-based planning domains, SBCO matches or exceeds a customized G\"odel Machine baseline while using $4$--$5.5\times$ less compute. 

Our analysis shows that SBCO improves precisely the constraints it can verify
reliably. It learns near-perfect, precision-heavy verifiers for constraints that
reduce to \emph{local, decidable predicates} (attribute lookups, thresholds,
presence checks, single-objective rankings) and abstains rather than mislead on
constraints that require a \emph{derived} signal (eg. geo-distance). These observations suggest that for constraint-based planning, the bottleneck is the \emph{structure of the improvement signal}: decomposed, correctness-grounded verification
is a far cheaper route to comparable or better quality than expensive
self-modification or metacognitive search\footnote{GenAI was used for editing and polishing the paper.}.


\bibliography{aaai2027}


\end{document}